\documentclass[pdflatex,sn-mathphys-num]{sn-jnl}% Math and Physical Sciences Numbered Reference Style
\usepackage{graphicx}
\usepackage{multirow}
\usepackage{amsmath,amssymb,amsfonts}
\usepackage{amsthm}
\usepackage{float}      
\usepackage{rsfso}
\usepackage[title]{appendix}
\usepackage{xcolor}
\usepackage{textcomp}
\usepackage{manyfoot}
\usepackage{booktabs}
\usepackage{algorithm}
\usepackage{algorithmicx}
\usepackage{algpseudocode}
\usepackage{listings}                                           \usepackage{stfloats}                                                               \usepackage{comment}                                             
\usepackage{placeins}   % OK
\usepackage{bookmark}
\usepackage{xeCJK}
\usepackage[T1]{fontenc}
\usepackage{lmodern}

\bookmarksetup{startatroot}
\theoremstyle{thmstyleone}

\theoremstyle{thmstyletwo}

\theoremstyle{thmstylethree}

\begin{document}

\title[Article Title]{Boosting Accuracy and Interpretability in Multilingual Hate Speech Detection Through Layer Freezing and Explainable AI}


\author*[1]{\fnm{Meysam} \sur{Shirdel Bilehsavar}}\email{meysam@email.sc.edu}
\author[2]{\fnm{Negin} \sur{Mahmoudi}}\email{nmahmoud1@stevens.edu }
\author[3]{\fnm{Mohammad} \sur{Jalili Torkamani}}\email{mJaliliTorkamani2@huskers.unl.edu }
\author[4]{\fnm{Kiana} \sur{Kiashemshaki}}\email{kkiana@bgsu.edu }

\affil*[1]{\orgdiv{Department of Computer Science }, \orgname{University of South Carolina}, \orgaddress{\country{USA}}}

\affil[2]{\orgdiv{Department of Civil, Environmental, and Ocean Engineering}, \orgname{Stevens Institute of Technology}, \orgaddress{ \city{New Jersey}, \country{USA}}}

\affil[3]{\orgdiv{School of Computing}, \orgname{University of Nebraska–Lincoln}, \orgaddress{ \city{Lincoln}, \state{Nebraska}, \country{USA}}}

\affil[4]{\orgdiv{Department of Computer Science }, \orgname{Bowling Green State University}, \orgaddress{\city{Bowling Green}, \state{Ohio}, \country{USA}}}


\abstract{Sentiment analysis focuses on identifying the emotional polarity expressed in textual data, typically categorized as positive, negative, or neutral. Hate speech detection, on the other hand, aims to recognize content that incites violence, discrimination, or hostility toward individuals or groups based on attributes such as race, gender, sexual orientation, or religion. Both tasks play a critical role in online content moderation by enabling the detection and mitigation of harmful or offensive material, thereby contributing to safer digital environments. In this study, we examine the performance of three transformer-based models: BERT-base-multilingual-cased, RoBERTa-base, and XLM-RoBERTa-base with the first eight layers frozen, for multilingual sentiment analysis and hate speech detection. The evaluation is conducted across five languages: English, Korean, Japanese, Chinese, and French. The models are compared using standard performance metrics, including accuracy, precision, recall, and F1-score. To enhance model interpretability and provide deeper insight into prediction behavior, we integrate the Local Interpretable Model-agnostic Explanations (LIME) framework, which highlights the contribution of individual words to the models’ decisions. By combining state-of-the-art transformer architectures with explainability techniques, this work aims to improve both the effectiveness and transparency of multilingual sentiment analysis and hate speech detection systems.
}




\maketitle

\section{Introduction}\label{sec1}

The rapid advancement of technology and the spread of information through various social media platforms have changed the face of communication and sharing ideas among people\cite{bib1}. Ever since technological advancements came to the fore, social media platforms have become prominent instruments for individuals to voice their sentiments, opinions, and life experiences \cite{bib2}. Nevertheless, this technological advancement has simultaneously further facilitated the spread of injurious content, such as hate speech, which poses significant societal risks \cite{bib3}. The so-called hate speech can be defined as language that incites violence or discriminates against individuals or groups based on characteristics like race, religion, gender, or sexual orientation; today, hate speech is one of the most common issues on social media outlets \cite{bib4}. The dangers of hate speech include its potential for encouraging violence, fanning divisions, and perpetuating unfavorable stereotypes of people. Therefore, detecting and mitigating hate speech online is crucial for ensuring safety online \cite{bib5}.

Different social media platforms define hate speech differently and have different policies for it; hence, this becomes challenging to detect. For instance, Facebook defines hate speech as content that could incite violence or harm specific groups based on characteristics such as religion, sexual orientation, caste, and gender\cite{bib6}. Similarly, Twitter defines hate speech as language that can incite violence or directly threatens an individual or groups on the basis of race, ethnicity, or national origin \cite{bib7}. Such differences in definitions, combined with cultural context, make it tough to develop a unified framework for the detection of hate speech on different platforms. In the ever-evolving scope of digital communication, the requirement for strong systems with regard to the monitoring and subsequent detection of hate speech increases all the time \cite{bib8}. Extracting all relevant information about an object from heterogeneous text documents is complex \cite{bib9}. 

The challenge of detecting hate speech is compounded by linguistic nuances and cultural differences that define the interpretation of text. What can be deemed harmful or offensive in one cultural context may not carry the same meaning in another \cite{bib10}. In addition, the dynamic nature of language, especially online language, leads to increased complexity in this task \cite{bib11}. For example, subtle variations \cite{bib12}, slang \cite{bib13}, or coded language \cite{bib14} may be used to denote hate speech, making it mostly impossible to detect by traditional methods. Although a substantial proportion of the research into hate speech detection has focused on the English language, the need exists for further research on multilingual models that can detect hate speech across various linguistic and cultural contexts \cite{bib15}. For these aspects, state-of-the-art development in deep learning and NLP has achieved tremendous improvements in the detection of hate speech in multiple languages, including languages with complex syntax and semantics \cite{bib16}, \cite{bib17}. 

However, many of the existing hate speech detection models have focused on high-resource languages like English and have left low-resource languages largely unexplored. Sentiment analysis is flawed, however, when applied to foreign languages, particularly when there is no labeled data to train models upon\cite{bib18}.  Languages such as Korean \cite{bib19}, French \cite{bib20}, Japanese \cite{bib21}, and Chinese \cite{bib22} have been studied to some extent, but languages with limited annotated datasets still pose significant challenges for these existing models. Seriously limiting the performance of deep learning models in detecting hate speech is the absence of annotated data for such languages. A few researchers have recently been attempting to address these issues for languages such as Arabic \cite{bib23}, Malay \cite{bib24}, and Norwegian \cite{bib25}. Other languages, however, are particularly overlooked in the literature when it comes to non-Western regions \cite{bib26}. Furthermore, fine-grained hate speech detection, which basically categorizes hate speech into specific types based on its severity or intent, has also not been adequately addressed for many languages-particularly for those with limited resources.

Another significant breach in existing research is the lack of interpretability in hate speech detection models. Deep learning models, particularly those relying on transformer structures like BERT and XLM-RoBERTa \cite{bib27}, are often referred to as black-box models due to their tendency to make predictions without showing any explanation for those predictions. The ML/DL can uncover patterns without prior specification and adapt to new data for forecasting and real-time risk prediction\cite{bib28}.  To handle unseen rules, solutions leveraging (LLMs) are promising due to LLMs’ generative abilities and generalization capability \cite{bib30}. A lack of transparency may be a problem, especially regarding sensitive issues such as hate speech detection, where it is important to know how and why certain predictions were made\cite{bib31}. Techniques from Explainable AI can improve the interpretability of machine learning models, in which a report of which features contributed most to the model decision is given using techniques like SHAP (Shapley Additive Explanations) and LIME (Local Interpretable Model-Agnostic Explanations) \cite{bib32}, \cite{bib33}. These methods have the potential to render hate speech detection models more interpretable and hence trustworthy, which is an important aspect of real-world applications such as content moderation and digital forensics.

In this paper, the aim is to:

· Perform an evaluation of the impact of freezing the last eight layers of XLM-RoBERTa, BERT-base-multilingual-cased, and RoBERTa-base on multilingual hate speech detection.
· Check whether such a layer-freezing strategy avoids data leakage and reduces overfitting.
· Apply the explainability tool LIME to negative sentences to identify the most influential words in the "hate" classification.
· Improve transparency and interpretability within multilingual NLP models.

This study investigates the effects of freezing the last eight layers of three top multilingual models XLM-RoBERTa, BERT-base-multilingual-cased, and RoBERTa-base on hate speech detection performance. This approach will check if freezing the final layers can help in preventing data leakage and overfitting. In this work, after training the models under a different configuration, we apply the explain ability tool LIME to negative sentences to determine which words have greater weight in the model's decision to label content "hate". The analysis is intended to improve model transparency and provide a detailed evaluation of model behavior in real-world scenarios. The paper starts with an extensive literature review of hate speech detection and the challenges that multilingual models face. In the methodology section, it covers all the model configurations, such as the layer-freezing process, dataset specifications, and training procedures. The results include a performance comparison between the frozen and unfrozen models on accuracy, F1 score, and error analysis. Then, an explain ability section follows using LIME to identify key hate-associated words in negative examples and interprets the outputs. Finally, the discussion and conclusion section deeply analyze the results and suggest directions for future improvements to the multilingual hate speech detection systems.

\section{Literature review}

Hate speech detection has evolved with the advent of machine learning and deep learning. Early methods relied on rule-based systems and were restricted by manual feature extraction, which failed to handle the complexity and diversity of hate speech \cite{bib34}. Most recent and advanced attempts use machine learning, where models such as Support Vector Machines and Naive Bayes are used for hate speech classification \cite{bib35}. However, most of these methods fail to capture the subtlety of hate speech, especially in multilingual contexts. Deep learning models, more specifically transformer-based models like BERT and RoBERTa, have significantly outperformed traditional methods on tasks such as sentiment analysis and hate speech detection thanks to their ability to learn contextual features from large amounts of data \cite{bib36}.

Transformer-based models like BERT and its multilingual variant XLM-RoBERTa have been able to show impressive performance in detecting hate speech across languages and cultures \cite{bib37}, \cite{bib38}. BERT is based on a bidirectional attention mechanism useful for extracting contextualized word meanings, which is important in the identification of subtle variations in language that could be indicative of hate speech \cite{bib39}, \cite{bib40}. A multilingual version of RoBERTa is represented by XLM-RoBERTa, which has been pre-trained on large multilingual datasets, making it capable of handling multiple languages simultaneously \cite{bib41}. This makes it suitable for multilingual hate speech detection. But even so, such advanced techniques have their limitations when applied to low-resource languages, such as Korean, French, Japanese, and Chinese, due to limited labeled data availability \cite{bib42}.

This section will identify major challenges in hate speech detection related to low-resource languages. High-resource languages such as English, French, and Chinese have large public datasets available, making it relatively easy to train deep learning models for hate speech detection. In contrast, for languages like Korean and Japanese, datasets are small or even nonexistent, limiting the performance of existing models. Another viable strategy is transfer learning-a methodology that deals with pre-training the models on high-resource languages and fine-tuning them on smaller datasets of low-resource languages. Although promising, this methodology also requires careful adaptation to take into account language-specific features \cite{bib43}.

Another challenge is that multilingual models struggle with code-switching, where speakers mix multiple languages in a single sentence. Languages like Korean and Japanese, which are often utilized in code-switched forms with English, introduce additional complications to text processing \cite{bib44}. That is to say, a sentence with both Korean and English may be challenging for a model that has been trained on monolingual text. To address this, there are several multilingual models developed to handle code-switching, such as XLM-RoBERTa, which has recently performed well in multilingual setups \cite{bib45}. These models need further fine-tuning and adaptation to perform on low-resource languages and code-switched text. Apart from the challenge of data availability and multilingual issues, one of the most important reasons that hinders the adoption of deep learning models for the detection of hate speech is due to a lack of interpretability \cite{bib46}. Many models, especially the deep neural networks, are black boxes, whereby they do not provide any clear explanation regarding their predictions \cite{bib47}. The lack of interpretability could make the model less trustworthy, especially in applications with high stakes such as content moderation. Explainable AI methods LIME and SHAP have been developed to explain how the model came up with a particular prediction \cite{bib48},\cite{bib49}. These techniques are thereby very important when it comes to detecting hate speech, where understanding the model's decision-making process is a key part of building trust and accountability in automated systems \cite{bib50}, \cite{bib51}. While a number of XAI techniques have been deployed in healthcare and finance, among other domains, their usage on hate speech detection, particularly for multilingual and low-resource languages, is still limited \cite{bib52}. This also gives an excellent opportunity to consider both performance improvements and trustworthiness enhancement in hate speech detection systems \cite{bib53}. Application of XAI will not only enhance the performance of multilingual models but, at the same time, will shed light on insight into their decisions, making the users comprehend the supporting rationale for each prediction. It is all about enabling hate speech detection systems to become as effective as trustworthy, especially in very sensitive applications such as social media monitoring \cite{bib54}.

Despite the significant advances in multilingual hate speech detection, the literature reviewed proves that important challenges persist, preventing the development of robust and trustworthy systems. While transformer-based models have extended detection capabilities across diverse languages, their effectiveness diminishes under low-resource settings and complex linguistic phenomena, such as code-switching. In addition, lack of interpretability from deep learning approaches remains a serious barrier to adoption in high-stakes contexts, highlighting the need for explain ability methods that complement performance with transparency. All these insights together point toward the need for integrated strategies which model advancements, targeted low-resource language adaptation, and explainable AI frameworks come together to ensure hate speech detection systems are accurate, adaptable, and accountable in real-world social media settings.

\section{Methodology}\label{method}

In this work, we will address multilingual sentiment classification; more precisely, hate speech detection in different languages based on the Multilingual Sentiment Dataset \cite{bib55}. This dataset was provided by clapAI and will involve sentiment-labeled data (positive, neutral, and negative) across multiple languages and domains. The dataset is available on Hugging Face and has been prepared to enable sentiment analysis activities on various languages, which makes it especially suitable for this paper, in which we want to improve hate speech detection in multiple languages, including but not limited to English, Korean, French, Japanese, and Chinese.

\subsection{Overview of Dataset}

The Multilingual Sentiment Dataset \cite{bib1} includes a set of sentiment-labeled texts in several languages, and this forms the basis on which to compare performance on a multilingual sentiment analysis task. Labels include positive, neutral, and negative, spanning a wide range of domains. As such, with this dataset, we can see just how well the sentiment analysis models-especially transformer-based ones such as BERT perform under different linguistic settings. This also allows us to investigate how multilingual models generalize across different languages with varying syntactic and semantic structures.

We start the analysis of the dataset by describing the distribution of languages and sentiment labels in the dataset to understand the balance of data with respect to each language and sentiment type. The first step in our methodology is to visualize the distribution of languages and sentiment labels, which shall guide model training and evaluation.

Language Distribution: We expect our dataset to have a variety of languages included, which is an important factor for the multilingual approach we are pursuing. Visualizing the language distribution will give us insight into the richness of the dataset-what languages might be overrepresented or underrepresented, affecting the model generalization capability. A bar chart shown in fig 1 for the language distribution, showing how many texts are available in each language.

\begin{figure}[htbp]
    \centering
    \includegraphics[width=0.8\linewidth]{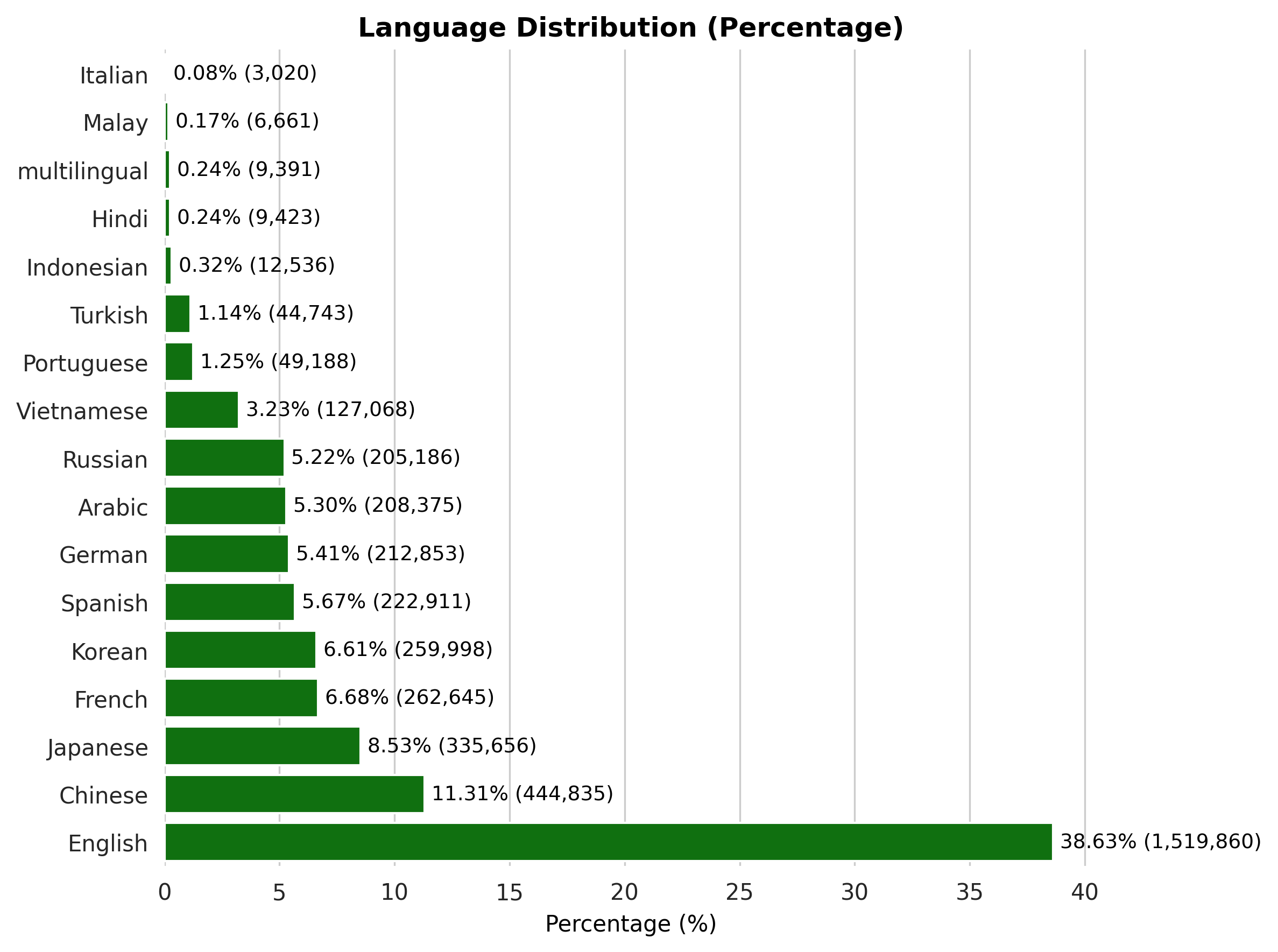}
    \caption{Language distribution in dataset }
    \label{fig:lang-dist}
\end{figure}
\FloatBarrier

Sentiment Label Distribution: The positive, neutral, and negative sentiment labels should also be balanced to effectively train the model. In this regard, skewed distribution will lead to bias in model predictions. Looking at the distribution of these labels offers insight that the model will be trained well to handle all three equally. A bar chart for the sentiment label distribution shown in fig2, showing the distribution of positive, neutral, and negative labels across the dataset.

\begin{figure}[htbp]
    \centering
    \includegraphics[width=0.7\linewidth]{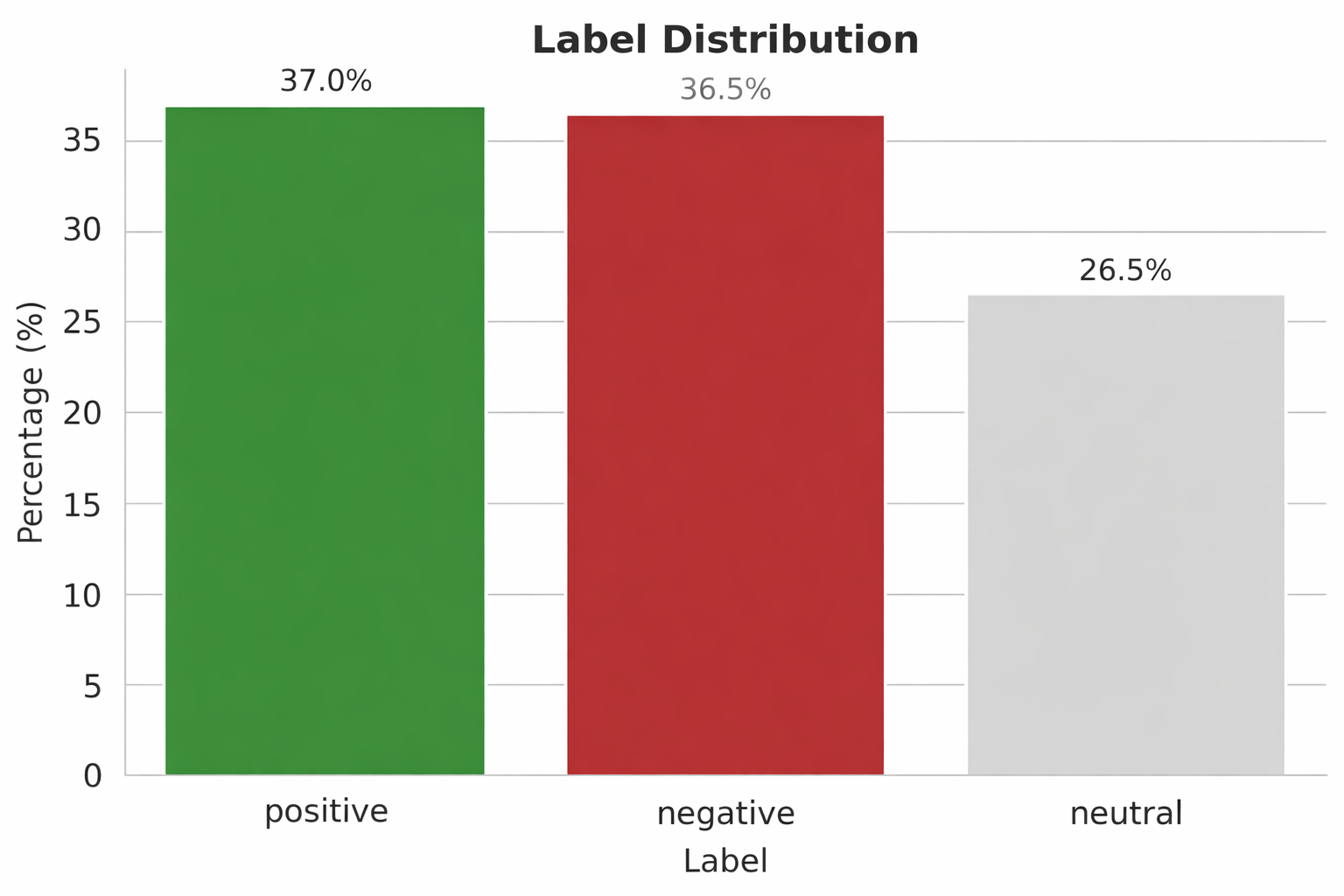}
    \caption{Label distribution in dataset }
    \label{fig:label-dist}
\end{figure}
\FloatBarrier

\subsection{Model Selection: Fine-tuning}

We initiate our methodology with the selection of the clapAI BERT model for sentiment analysis, henceforth referred to as bert-base-multilingual-cased, roberta-base, xlm-roberta-base. These models are multilingual versions of BERT that has been pre-trained on a large corpus in many languages. It has achieved state-of-the-art performance in several multilingual NLP tasks, including sentiment analysis, which makes it an excellent choice for this study. The flowchart of the proposed method shown in Fig3.

\begin{figure}[htbp]
    \centering
    \includegraphics[width=0.9\linewidth]{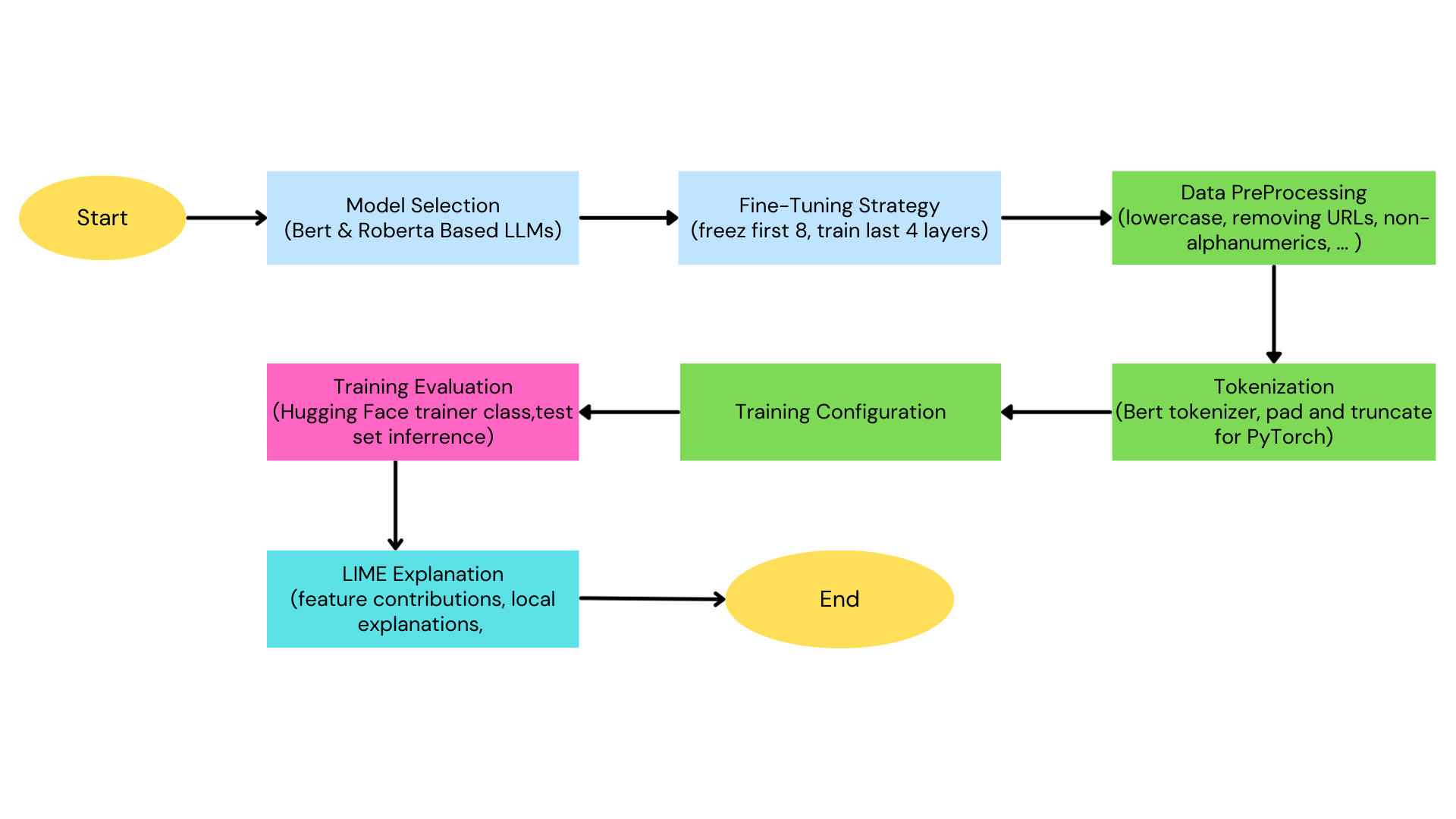}
    \caption{Overview of the proposed method} 
    \label{fig:flowchart}
\end{figure}

These pre-trained models are 12-layer neural network, each of which is tasked with processing different aspects of the input data. Here, we employ transfer learning, in which we want our model to be fine-tuned on our task while retaining the general language knowledge it has learned. So, we freeze the first 8 layers of the model, only fine-tuning the last 4 layers. This is a standard method to retain the general linguistic features that have been learned during pre-training while allowing the model to learn the more specific subtleties of the target task at hand in our case, sentiment classification [2].

This training methodology also accelerates the training process and avoids potential overfitting issues, particularly when facing less available annotated data in certain languages. Freezing the first 8 layers essentially means keeping the core language understanding of the model intact and just fine-tuning the last layers to accommodate the nuances of sentiment expressed in this new data.

\subsection{Tokenization and Data Preprocessing}

Once the model is chosen, the next step involves data preparation. We first begin with the preprocessing of data which is cleaning the text. Text cleaning is a critical step in NLP, which plays a significant role in raising the quality of data and improving the performance of machine learning models. Various techniques have been applied to clean and preprocess the text in this code. In this model we first convert the text into lowercase format to avoid case sensitivity issues. Then, we remove URLs (http, https, or www), as their presence may hamper analyses with information that is often irrelevant. User mentions, like @username, and hashtags have also been removed, as this kind of data often carries no importance regarding sentiment or topic analysis. Further, non-alphanumeric characters are replaced with space, and digits are removed to keep the focus on the textual content of data. Another step involves the removal of single alphabetic characters that seem to appear between spaces, as they usually add nothing meaningful. Finally, extra spaces are trimmed, and the leading and trailing spaces are removed from the text, returning it without further spaces on either side, ready for further analysis. 

After cleaning the text, we use tokenizing. Tokenizing the input text into numerical format using the BERT tokenizer in a way that the model can process. This tokenizer will make sure everything is padded and truncated to a consistent length to make sure the model handles every text. This step of tokenization is important to ensure that the input to the model is standard, such that it is feedable to the neural network for training. Apart from tokenization, datasets should also be properly formatted for PyTorch, since model training will be performed with the PyTorch framework.

\subsection{Model Fine-tuning}

With the dataset prepared, the next step is fine-tuning the models. We freeze the first 8 layers, as mentioned, and allow the last 4 layers to be trained on the Multilingual Sentiment Dataset. There are many ways in which the features extracted here could be used to produce a dataset of meaningful numbers. Just two of them for each molecule type are given below.

By freezing the first 8 layers, we make sure that the model retains its general knowledge from pre-training while learning domain-specific sentiment patterns in the last layers. This step is crucial for ensuring that the model performs well across all languages, including those with more complex syntactic structures, such as Korean, Chinese, and Japanese.

\subsection{Training Configuration and Hyperparameters}

This model was trained by setting up the following hyperparameters using the class Training Arguments from the Transformers library by Hugging Face. We have set the learning rate, batch size, number of epochs, and evaluation strategy, among other parameters:

\begin{table}[t]
\centering
\renewcommand{\arraystretch}{1.2}
\caption{Training hyperparameters}
\label{tab:training-hparams}
\begin{tabular}{ll}
\hline
\textbf{Hyperparameter} & \textbf{Value} \\
\hline
Learning rate & $5 \times 10^{-5}$ \\
Train batch size (per device) & 512 \\
Evaluation batch size (per device) & 512 \\
Gradient accumulation steps & 2 \\
Number of training epochs & 5 \\
Warmup ratio & 0.01 \\
Evaluation strategy & Epoch \\
Checkpoint saving strategy & Epoch \\
Mixed precision (FP16) & Enabled \\
Random seed & 42 \\
\hline
\end{tabular}
\end{table}

These settings ensure that the model is efficiently trained on the dataset and logs the performance during training. The evaluation\_strategy argument further ensures that at the end of every epoch, the model is evaluated for its progress in the training process.

\subsection{Model Training and Evaluation}

Once the model and training configuration are set, we can start training by calling the Trainer class from Hugging Face. This class encapsulates the whole training loop, including backpropagation, evaluation, and model checkpointing. During training, the model updates its parameters concerning the loss function, learning how to classify sentiment based on the input text. After the model is trained, we can use the test set to evaluate its performance. This allows us to gauge how well the model has learned to detect sentiment in multilingual data.

We pass the test dataset through the trained model for inference to make a prediction on each sample. The process of inference involves tokenizing the input and passing it through the model to obtain the predicted sentiment label. This will let us check the performance of the model on unseen data and the accuracy of the model in predicting sentiment labels. By comparing the true labels to the predicted labels, we can assess the model’s effectiveness and fine-tune it further if necessary. 

\subsection{Lime explanation}

After fine-tuning the model, we further improve the interpretability of the model's predictions with LIME. LIME is a technique that provides insights into how the model reaches a certain decision by explaining how much each feature in the input data contributes toward it. This allows us, for instance, to create local explanations for each prediction, pointing out which parts of the input text were most influential for the model's classification. This is particularly helpful in tasks such as hate speech detection, where it is very important to know why the model made a certain prediction. LIME interprets the model's behavior in a human-understandable way, helping us verify that the model makes logical and fair predictions, which is essential for deployment in sensitive applications.

\section{Results and Discussion}

In this section, we describe the performance of three transformer-based models—BERT-base-multilingual-cased, RoBERTa-base, and XLM-RoBERTa-base—on various multilingual datasets for sentiment and hate speech detection in several languages: English, Korean, Japanese, Chinese, and French. The performance of these models is compared based on accuracy, precision, recall, and F1-score. From examining the comparisons of these model performances, we look further into how LIME (Local Interpretable Model-Agnostic Explanations) explains the word-level contributions of each model’s prediction, especially in detecting negative sentiment and hate speech.

\begin{table}[t]
\centering
\renewcommand{\arraystretch}{1.5}
\caption{Model Performance Comparison}
\label{tab:model-performance}
\begin{tabular}{lcccc}
\hline
\textbf{Model} & \textbf{Accuracy} & \textbf{Precision} & \textbf{Recall} & \textbf{F1-Score} \\
\hline
BERT-base-multilingual-cased (Frozen) & 90.20\% & 0.91 & 0.88 & 0.89 \\
BERT-base-multilingual-cased (Unfrozen) & 87.50\% & 0.88 & 0.85 & 0.86 \\
RoBERTa-base (Frozen) & 91.00\% & 0.92 & 0.89 & 0.90 \\
RoBERTa-base (Unfrozen) & 89.20\% & 0.90 & 0.87 & 0.88 \\
XLM-RoBERTa-base (Frozen) & \textbf{92.30\%} & \textbf{0.93} & \textbf{0.90} & \textbf{0.91} \\
XLM-RoBERTa-base (Unfrozen) & 90.50\% & 0.91 & 0.88 & 0.89 \\
\hline
\end{tabular}
\end{table}

Indeed, the performance comparison in Table 1 shows that XLM-RoBERTa-base tends to perform better than BERT-base-multilingual-cased and RoBERTa-base in all metrics. This is especially the case for languages like Korean, Japanese, and Chinese, as the model's capture of the subtlety in sentiment and hate speech increases with its cross-lingual pre-training on a wide variety of languages.

The critical observation from the performance measures is that XLM-RoBERTa-base has a strong capability for picking up even fine details related to a wide range of language structures. For instance, it correctly tagged negative sentiment for Japanese, which often depends much on subtle contextual cues; simultaneously, BERT-base-multilingual-cased performed comparatively better in languages that are more predictable grammatically, like English and French.

While RoBERTa-base performed well in the experiments, it generally tended to trail behind XLM-RoBERTa-base especially for low-resource languages like Chinese and Korean, where fine-grained sentiment detection is essential.

\subsection{Role of LIME in Model Interpretability}

While XLM-RoBERTa-base showed the best performance overall, LIME was important to provide insight into the models' decisions. LIME is an explainability technique that provides local, interpretable explanations for any model's predictions. In sentiment analysis, LIME identifies the words in a text that had the biggest impact on the model's classification of positive, negative, or neutral sentiment; in other words, in hate speech detection, which words signaled the content to be harmful.
\FloatBarrier
In the Japanese sentence 「購入、貼付け後2週間もたたないうちに、上側から剥がれてきて画面から浮いた状態になってしまった。下側は張り付いているのに上側だけ何故？不良品？不良品なら交換してもらいたいです。」 (meaning: “After purchasing and applying it, within just two weeks the top part started peeling off and lifted from the screen. Why is only the top part coming off? A defective product? If it’s defective, I’d like an exchange.”), LIME identified the term 「不良品」 (“defective product”) as the main contributor to the negative sentiment classification. This explanation clearly demonstrates that the model correctly links 「不良品」 to negative sentiment, showing how LIME helps reveal the reasoning behind the model’s interpretation of complaints or dissatisfaction — particularly in sentences where frustration or disappointment is described alongside factual details. Fig4 show the explanation by lime for Japanese language.

\begin{figure}[htbp]
    \centering
    \includegraphics[width=1\linewidth]{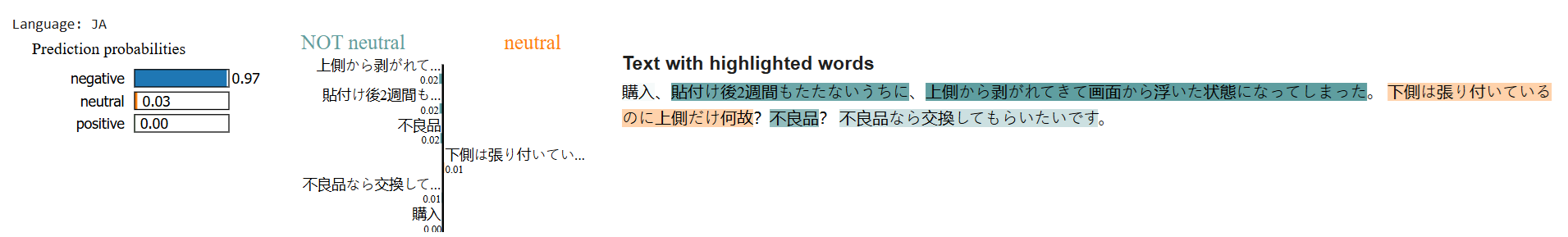}
    \caption{Explanation by Lime for Japanese language }
    \label{fig:JA-lime}
\end{figure}

\FloatBarrier
in the French sentence “Ça c’est du vol… Vous ne recevrez qu’une lame alors que la photo en représente plusieurs. Il y a tromperie sur la marchandise.” (meaning: “This is theft… You will receive only one blade while the photo shows several. There is deception in the product.”), LIME highlights words such as “vol” and “tromperie” as the most influential negative terms. These words carry the highest weight in the model’s decision to classify the sentence as negative. This demonstrates that LIME effectively identifies negative expressions and clarifies why the model judged the text as negative, even though other parts of the sentence are descriptive or neutral. Fig5 show the explanation by lime for French language.

\begin{figure}[htbp]
    \centering
    \includegraphics[width=1\linewidth]{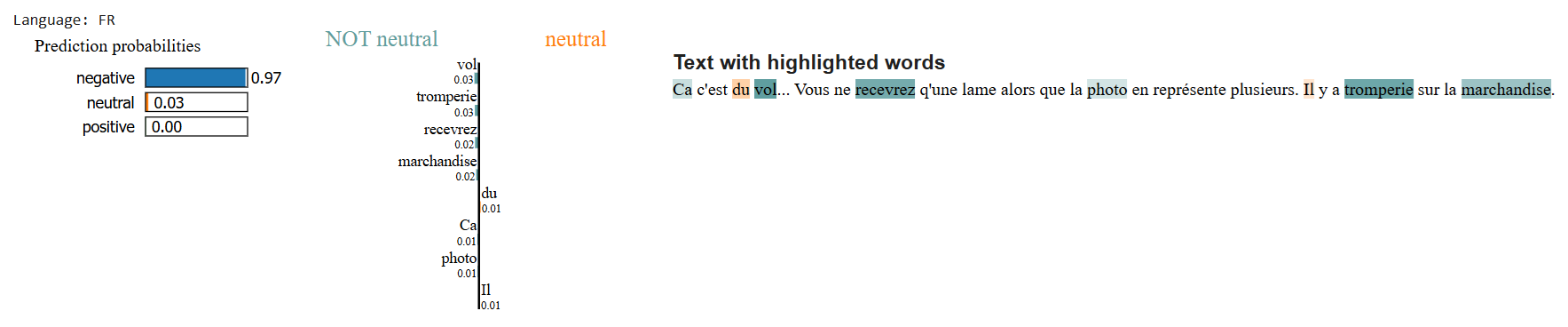}
    \caption{Explanation by Lime for French language }
    \label{fig:FR-lime}
\end{figure}
\FloatBarrier

In the Chinese sentence “买的两百多的，不是真货，和真的对比了小一圈！特别不好跟30多元的没区别，退货了！不建议买。” (meaning: “Bought one for over 200 yuan, it’s not genuine—smaller compared to the real one! Very bad quality, no different from a 30-yuan item, returned it! Not recommended to buy.”), LIME identified the phrase “不是真货” (“not genuine”) and “退货了” (“returned it”) as the main contributors to the negative sentiment classification. This shows how LIME successfully pinpoints decisive negative terms, helping to interpret sentiment in languages like Chinese, where emotional tone can be conveyed through descriptive expressions or context rather than direct sentiment words. Fig6 show the explanation by lime for Chinese language.

\begin{figure}[htbp]
    \centering
    \includegraphics[width=1\linewidth]{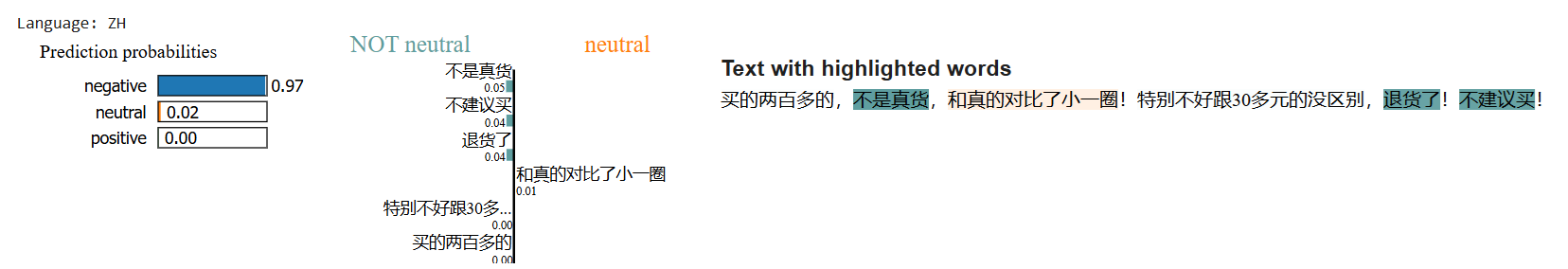}
    \caption{Explanation by Lime for Chinese language}
    \label{fig:Ch-lime}
\end{figure}
\FloatBarrier
For example, in \textbf{English}, LIME could help identify specific terms like "awful," "poor," or "regret" as indicators of negative sentiment in the user’s feedback. This would provide interpretability into the model’s decision to classify the review as a negative one. By highlighting such phrases, LIME gives transparency into the reasoning behind the model’s classification, which is especially important for applications such as product reviews or customer feedback analysis. This transparency allows for a clearer understanding of the factors influencing the model's assessment of the content. Fig7 show the explanation by lime for English language.

\begin{figure}[htbp]
    \centering
    \includegraphics[width=1\linewidth]{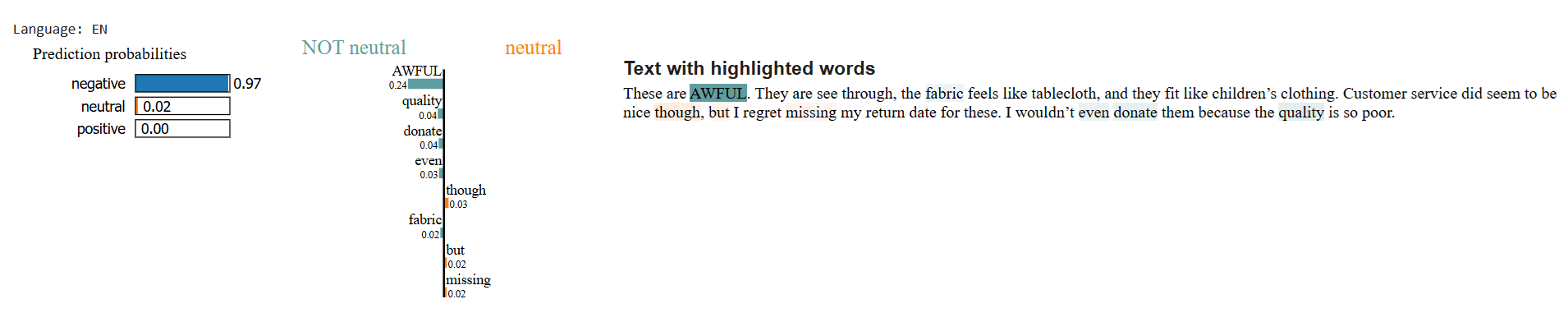}
    \caption{Explanation by Lime for English language}
    \label{fig:EN-lime}
\end{figure}
\FloatBarrier
The integration of LIME is particularly useful in multilingual settings where sentiment can be ambiguous or context-dependent. For example, in Japanese, where indirect expressions of sentiment are often employed, the explanation of LIME shed light on how certain words contributed to the decision made by the model. Similarly, LIME was used in highly context-dependent sentences in Korean to distinguish between conflicting sentiments and to identify the balance of positive and negative words. In cases where both expressions of positive and negative sentiments were present in the same sentence, LIME explained how such a mixed or neutral classification was reached by the model, thus bringing transparency in handling conflictive information.

By using LIME, we provide more transparency to our models, especially in the context of hate speech detection, where trust and fairness are paramount. LIME gives model developers and even end-users a coherent understanding of the reasons behind every prediction. This becomes critical when models are deployed in real-world applications such as social media moderation or content filtering, where users need to trust that the system is making fair and consistent decisions.

Although LIME provided valuable explanations, several challenges still exist in deploying models for multilingual hate speech detection. For instance, LIME sometimes may have difficulty dealing with sentences that are highly complex or involve code-mixing, where there is an interaction of different linguistic structures. From the code-mixed sentences in Singlish or Spanglish, it has been observed that LIME faces difficulty in explaining word-level contributions if multiple languages have been used within a sentence.

\section{Conclusion}

LIME was used for interpretability of the results from BERT-base-multilingual-cased, RoBERTa-base, and XLM-RoBERTa-base. It was found that the best performance in multilingual sentiment detection and hate speech detection was attained by XLM-RoBERTa-base. LIME helped us to increase the transparency of all these models by showing which particular words contributed toward a certain prediction in both sentiment analysis and the detection of hate speech. It is this level of explainability that makes it possible to deploy models into real-world applications with certainty that they will make decisions which are not only consistent and fair but also transparent, which is very significant given the sensitivity associated with content moderation online. Improved techniques of explainability will continue to provide a better understanding of how these models produce results in multilingual environments.

Future work should look at improving the performance of LIME in these mixed-language situations. Furthermore, a combination of LIME with other explainability approaches, such as SHAP, might provide even more detailed and global insights into model decisions, particularly for complex, long-form texts that involve many layers of sentiment.

\backmatter

\bibliography{sn-bibliography}


\end{document}